# Sexism Identification in Tweets and Gabs using Deep Neural Networks

Amikul Kalra
{a.kalra@se20.qmul.ac.uk}

Arkaitz Zubiaga
{a.zubiaga@qmul.ac.uk}

School of Electronic Engineering and Computer Science,
Queen Mary Universiy of London

*Abstract*— Through anonymisation and accessibility, social media platforms have facilitated the proliferation of hate speech, prompting increased research in developing automatic methods to identify these texts. This paper explores the classification of sexism in text using a variety of deep neural network model architectures such as Long-Short-Term Memory (LSTMs) and Convolutional Neural Networks (CNNs). These networks are used in conjunction with transfer learning in the form of Bidirectional Encoder Representations from Transformers (BERT) and DistilBERT models, along with data augmentation, to perform binary and multiclass sexism classification on the dataset of tweets and gabs from the sEXism Identification in Social neTworks (EXIST) task in IberLEF 2021. The models are seen to perform comparatively to those from the competition, with the best performances seen using BERT and a multi-filter CNN model. Data augmentation further improves these results for the multi-class classification task. This paper also explores the errors made by the models and discusses the difficulty in automatically classifying sexism due to the subjectivity of the labels and the complexity of natural language used in social media.

## I. Introduction

Social media has completely altered the way communities are formed and utilised, which provides incredible advantages while also having severe repercussions. Due to the 'online disinhibition effect' (Suler, 2004, p. 1), when users are provided with an anonymised and accessible platform, they engage in behaviours they would not partake in when interacting face-to-face (Wright et al., 2019). A significant example of this is the hate speech produced and propagated through social media platforms. Hate speech is defined as language which is 'insulting, degrading, defaming, negatively stereotyping, or inciting hatred, discrimination or violence against people in virtue of their race, ethnicity, nationality, religion, sexual orientation, disability, gender identity' (Brown, 2017, p. 1). The prevalence of hate speech in everyday life has increased in correlation with social media usage, particularly during the COVID-19 pandemic, with internet usage levels having increased between 50% to 70% as of early April 2020 (UN Women, 2020). Online hate speech, especially targeted discrimination, has been associated with an increase in hate crimes offline (Hatzipanagos, 2018; Laub, 2019; Relia et al., 2019); therefore, the ability to successfully tackle this issue within the virtual space itself is vital.

Sexism refers to a sub-classification of hate speech where the targeted people are typically female. Women are more likely to report having experienced sexual harassment online (16% vs. 5%) or being cyber-stalked (13% vs. 9%) compared to men (Vogels, 2021); with 1 in 10 women reporting having experienced cyber harassment since the age of 15 in the European Union (UN Women, 2020). Women and girls are specifically seen to face a digital gender divide[1], especially with the COVID-19 pandemic being the first major one in the age of social media[2].

While social media platforms like Twitter do ban hate speech[3], these policies are enforced primarily through manual methods which cannot scale up to counteract the data being produced (Waseem and Hovy, 2016; Zhang and Luo, 2019). Hence, more automatic methods are essential to successfully tackle this problem. This paper looks at creating and improving neural network models to perform binary and multiclass sexism classification using the dataset provided for the first shared task on sEXism Identification in Social neTworks (EXIST) at IberLEF 2021 (Rodríguez-Sánchez et al., 2021). In this paper, sexism refers to hate speech against women specifically, with the dataset consisting of texts obtained from Twitter and Gab. This paper presents a variety of models including Long-Short-Term Memory networks (LSTMs), Bidirectional Long-Short-Term Memory networks (Bi-LSTMs), and Convolutional Neural Networks (CNNs) tested against this dataset, with the best performance for both tasks seen with a model utilising Bidirectional Encoder Representations from Transformers (BERT) contextual embeddings in conjunction with a CNN containing three filter sizes of 4, 6, and 8. Data augmentation is also seen to improve the model's performance for the multi-class classification task. The performance metrics are reported for each model with a discussion of the errors presented to explore the difficulty in classifying a subjective topic like sexism in natural language texts.

The rest of this paper is organised as follows. In Section 2, related work in the field of hate speech and sexism classification are looked at, Section 3 talks about the dataset used for the experiments, Section 4 talks about system architecture and details the models used in this paper, Section 5 presents and discusses the results, Section 6 talks about error analysis and Section 7 concludes the paper.

---

[1] https://itu.foleon.com/itu/measuring-digital-development/gender-gap/
[2] https://www.vox.com/recode/2020/3/12/21175570/coronavirus-covid-19-social-media-twitter-facebook-google
[3] https://help.twitter.com/en/rules-and-policies/hateful-conduct-policy

## II. RELATED WORK

Research has been conducted regarding hate speech on various social media platforms like Twitter (Davidson et al., 2017; Shushkevich and Cardiff, 2018; Rodríguez-Sánchez et al., 2020), Facebook, (Vigna et al., 2017; Raiyani et al., 2018; Mandl et al., 2019) and Reddit (Qian et al., 2019). Initial research regarding classifying hate speech was feature based, with approaches based on bag-of-words (BoW), character-level, and word-level n-grams (Kwok and Wang, 2013; Mehdad and Tetreault, 2016; Waseem and Hovy, 2016). Then, traditional machine learning methods like support vector machines (SVM) and logistic regression were used. One of the first uses of machine learning to detect offensive tweets was presented by Xiang et al. (2012) where logistic regression was used as opposed to the utilisation of pattern-based approaches (Gianfortoni et al., 2011; Mondal et al., 2017).

One of the first instances of using neural networks and word embeddings to tackle hate speech classification was done by Djuric et al. (2015) where a continuous BoW model was used to learn paragraph2vec embeddings, with these embeddings then used to train a binary classifier.

For the 2018 IberEval Automatic Misogyny Identification (AMI) tasks, which required classification of English and Spanish or English and Italian tweets, the majority of the participants utilised SVMs and Ensemble of Classifiers (EoC) (Ahluwalia et al., 2018; Fersini et al., 2018; Pamungkas et al., 2018; Shushkevich and Cardiff, 2018). Good results were also seen through the usage of Bi-LSTMs and Conditional Random Fields (CRFs) on the same task (Goenaga et al., 2018). An alternative architecture was proposed by Zhang and Luo (2019) with two deep neural network models to tackle hate speech classification on Twitter datasets. These models consist of CNN and Gated Recurrent Unit (GRU) architectures with the results outperforming the best methods at the time.

More recently, the introduction of BERT has led to new state-of-the-art performances across a range of natural language processing tasks, including text classification (Devlin et al., 2018). BERT is a multi-layer bidirectional transformer encoder which notably uses bidirectional self-attention to learn contextual information between words and sub-words within a text (Alammar, 2018). BERT has been pre-trained using BooksCorpus (800M words) and English Wikipedia (2500M words) on masked language modelling and next sentence prediction (Devlin et al., 2018). This causes the embeddings taken from the model to contain useful contextual information that can be fine-tuned for specific tasks. Rodríguez-Sánchez et al. (2020) show BERT being used to give the best performance on the task of identifying sexist content through fine tuning pre-trained mBERT-Base parameters with a fully connected layer.

Multi-label sexism classification was first seen in a paper by Parikh et al. (2020) where a BERT based neural architecture was used along with distributional and word level embeddings. Samghabadi et al. (2020) also show BERT being used without fine-tuning to identify aggression and misogyny in English, Hindi, and Bengali tweets with positive results from the model.

Limited research has been conducted on the automatic classification of subtle expressions of sexism encompassing a broad range of categories, compared to the sole use of profanities or explicit hatred against women. Rodríguez-Sánchez et al. (2020) collected instances of various types of sexism, ranging from subtle inequality to explicit violence to create a dataset to then be used in an automatic classification task. The range of expressions collated is similar to the dataset used in this paper. An important point to note is that hate speech and sexism is defined differently across these papers, with offensive language often considered to be equivalent (Davidson et al., 2017). Another challenge for automatic sexism classification is the lack of an established benchmark dataset.

Detecting sexist expressions is a challenge for human coders as well, with racist or homophobic tweets often considered to be hate speech while sexist or derogatory terms are found to be offensive as opposed to hateful (Waseem and Hovy, 2016; Davidson et al., 2017). To tackle these issues of subtlety and context, the DistilBERT, which is a lightweight version of BERT with 40% fewer parameters (Sanh et al., 2019), and BERT models used in this paper are fine-tuned with additional layers to allow them to learn contextual embeddings and perform effectively on the sexism classification tasks.

## III. DATASET

The dataset was taken from the first shared task at IberLEF EXIST 2021[4]. It originally contained English and Spanish tweets and gabs with solely the English texts used in this paper. The training dataset contains 3436 tweets and the test set consists of 1716 tweets and 492 gabs. For the first classification task, each text was labelled as 'sexist' or 'non-sexist' while the second task further categorised the same texts into one of the following categories: (i) 'ideological-inequality', (ii) 'objectification', (iii) 'sexual-violence', (iv) 'stereotyping-dominance', (v) 'misogyny-non-sexual-violence', (vi) 'non-sexist'. The breakdown of classes in the English dataset can be seen in Table I. The training dataset was split to use 80% of the data for training and 20% for validation for each model with no shuffling, with the reported results observed through testing it on the test dataset of 2208 texts.

TABLE I.
BREAKDOWN OF THE TRAINING AND TEST DATASET FOR BINARY CLASSIFICATION (TASK 1) AND MULTICLASS CLASSIFICATION (TASK 2).

| Task 1 | Train | Test | Task 2 | Train | Test | Examples (no pre-processing) |
|---|---|---|---|---|---|---|
| sexist | 48% | 52% | ideological-inequality | 11% | 15% | 'I think the whole equality thing is getting out of hand. We are different, thats how were made!' |
| | | | objectification | 7% | 7% | '@user @user Wow, your skirt is very short. What is it's length? 5 inch or more?' |
| | | | sexual-violence | 10% | 9% | 'B*tches be begging me to fw them just to give me a reason not to fw them. Lol' |
| | | | stereotyping-dominance | 11% | 12% | 'well yeah ofcourse you'll cook for him he'll be your husband after all URL' |
| | | | misogyny-non-sexual-violence | 8% | 10% | '@user This why I hate women' |
| non-sexist | 52% | 48% | non-sexist | 52% | 48% | '@user This is a super news for the #WomensRights.' |

---
[4] http://nlp.uned.es/exist2021/

The following pre-processing steps were used on the training and test datasets:
- Hyphens ('-') were replaced with a space to split up the hyphenated terms.
- Hashtag symbols ('#') were replaced with a space while the hashtag terms themselves were retained without the symbol as these often indicate topics of conversation and would be helpful as features.
- All other punctuation, except apostrophes, was removed (@"!*^&()%$,.:;[]{}=~_+?\|/"")
- Stopwords were not removed as these could impact the sentiment of a tweet, for example, negation through the word 'not'.
- All terms were lowercased before being used in the models.
- Twitter and Gab usernames were replaced with the term 'username'.
- URLs were removed from the texts

## IV. SYSTEM ARCHITECTURE

The Keras library is used for all the models in this paper with the hugging-transformers library [5] used for the pre-trained BERT and DistilBERT models. The DistilBERT model used is 'distilbert-base-uncased' while the BERT model is 'bert-base-uncased'. Both these models were initialised with a dropout and attention dropout rate of 0.2. The baseline system for each task in the competition was an SVM classifier trained on unigram representations of the texts as bag-of-words (Rodríguez-Sánchez et al., 2021).

### A. Models

The first model used is a simple bag of words where word embeddings of dimension 100 were learned through training. A variation of this bag of words model used Global Vectors for Word Representation (GloVE) embeddings (Pennington et al., 2014) to observe the impact of using pre-trained vectors both with and without fine-tuning the learned weights.

DistilBERT was then utilised in a variety of architectures to explore the impact on model performance. The first variant was the pre-trained model for sequence classification from the transformers library with no modification. Following this, the DistilBERT model was used to obtain contextual word embeddings for the text, with the sentence embedding then fed through a fully connected layer to provide a classification decision.

Experimentation was then done with the embedding output from the DistilBERT model in conjunction with the following more complex architectures prior to the final fully connected layer: (i) an LSTM layer, (ii) a Bi-LSTM layer,

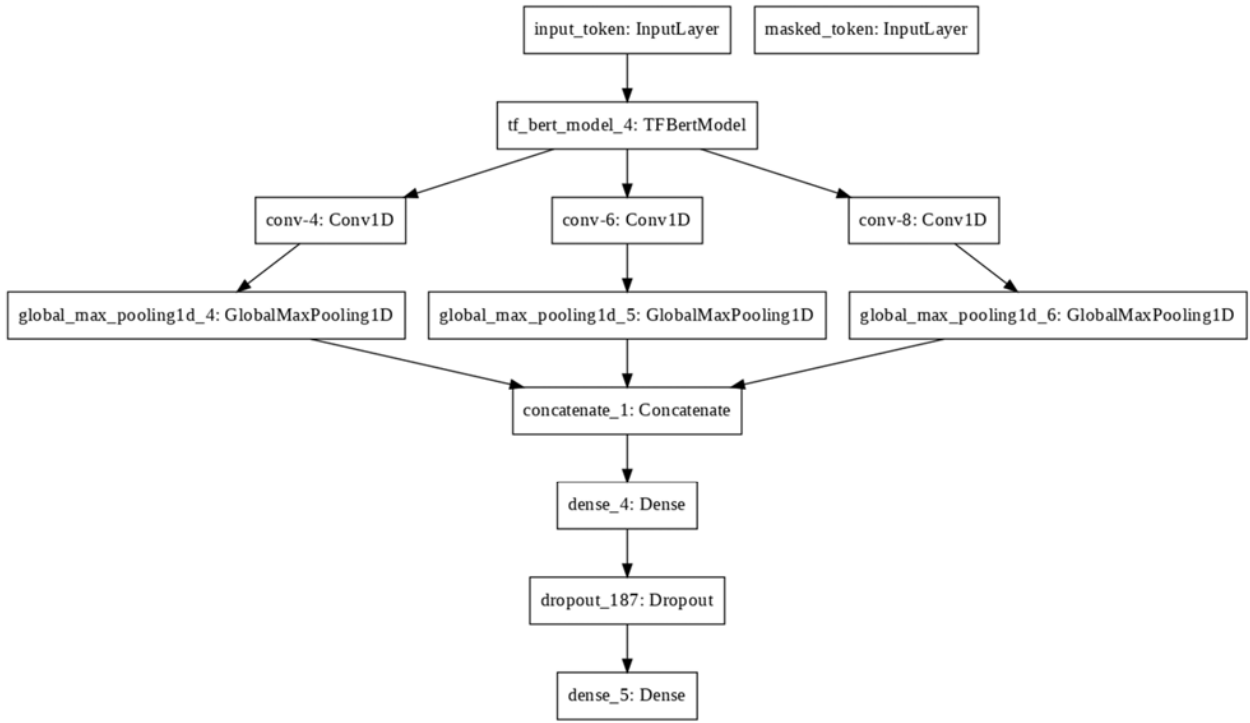

Fig. 1. The structure of the MultiCNN model used in conjunction with the BERT embeddings. The dropout layer rate is set to 0.2 and the embedding size is 100.

---

[5] https://huggingface.co/transformers/index.html

(iii) a CNN layer. The two variations of CNN architectures used in this paper are: (i) a single 1D convolutional layer with a filter size of 6 and (ii) 3 convolutional filters of size 4, 6, and 8 respectively. The different sized filters would then capture information from 4, 6, and 8-gram sub-sequences within the text, with the intention of being able to capture sentiments more clearly as opposed to an architecture using a recurrent neural network like an LSTM. The outputs from these three filters were then max pooled before being fed through two fully connected layers with a dropout layer in between. A visualisation of this model, called MultiCNN in this paper, can be seen in Fig. 1. Further experimentation involved the CNN models being modified by replacing the DistilBERT embeddings with BERT embeddings to improve the results.

The same models were used for Task 1 and Task 2, with the only difference being the use of a sigmoid activation function for the binary classification task with the loss calculated as 'binary cross entropy', while the second task used a softmax activation function with 'categorical cross entropy' for the loss function. A dropout rate of 0.2 was used for all the models, except where specified otherwise; and the Adam optimiser was used while training with the learning rate set to $5 \times 10^{-5}$. All the models were run for 50 epochs with early stopping implemented and the best weights returned if the accuracy did not improve for 15 epochs. The reported results have been averaged over 5 runs.

### B. Data Augmentation

The other major contribution from this paper was the use of data augmentation on the training dataset to improve the overfitting seen with the BERT and DistilBERT models, by simply providing more training data for the model to utilise. This paper utilised Easy Data Augmentation (EDA) introduced by Zou and Wei (2019) for this purpose. The augmentation consists of four operations: synonym replacement, random insertion, random swap, and random deletion. These operations, apart from random deletion, were performed on the pre-processed training dataset to produce 8 augmented sentences for each original sentence. The augmented sentences were annotated with the same label as the original sentence. Deletion was not used as it could potentially remove a term that causes the sentence to be labelled as sexist or non-sexist in this dataset, and therefore the label for the augmented sentence would then not be accurate. A rate 0.05 was set for all the operations, which refers to the percentage of words affected in a sentence with the augmentation operations. This rate of 0.05, along with the use of 8 augmented sentences per original, was taken from the recommendations in the EDA paper in accordance with the size of the training data set used (Zou and Wei, 2019).

For both tasks, data augmentation was used in combination with the BERT + CNN and BERT + MultiCNN models as these were achieving the highest accuracy and F1 scores across the models. For the second task, a variation of this data augmented model utilised class weight information as an additional input to investigate the impact on the F1 score through addressing the imbalanced classes.

## V. RESULTS

The averaged results for Task 1 can be seen in Table II where the models had to make a binary classification decision and label the text as 'sexist' or 'non-sexist'. The results for the simple BoW models with no pre-trained embeddings or

TABLE II.
TASK 1 RESULTS

| Model | Accuracy | F1 Score | Precision | Recall |
|---|---|---|---|---|
| BoW model | 0.542 | 0.415 | 0.535 | 0.403 |
| GLoVE embeddings | 0.474 | 0.410 | 0.488 | 0.382 |
| GLoVE embeddings (finetuned) | 0.380 | 0.258 | 0.399 | 0.219 |
| TFDistilBertForSequenceClassification | 0.725 | | | |
| DistilBERT NBoW (no dropout) | 0.737 | 0.736 | 0.736 | 0.738 |
| DistilBERT NBoW | 0.739 | 0.737 | 0.737 | 0.740 |
| DistilBERT + LSTM | 0.745 | 0.743 | 0.743 | 0.746 |
| DistilBERT + Bidirectional LSTM | 0.749 | 0.747 | 0.747 | 0.750 |
| DistilBERT + CNN | 0.745 | 0.744 | 0.745 | 0.745 |
| DistilBERT + MultiCNN | 0.747 | 0.746 | 0.746 | 0.747 |
| BERT + CNN | 0.752 | 0.751 | 0.751 | 0.751 |
| BERT + CNN + Dropout(0.4) | 0.748 | 0.747 | 0.747 | 0.748 |
| **BERT + MultiCNN** | **0.762** | **0.760** | **0.760** | **0.764** |
| BERT + MultiCNN + Dropout(0.4) | 0.759 | 0.757 | 0.757 | 0.762 |
| BERT + CNN + Data Augmentation | 0.748 | 0.747 | 0.747 | 0.748 |
| BERT + MultiCNN + Data Augmentation | 0.749 | 0.748 | 0.748 | 0.749 |
| **Competition Results (Rodríguez-Sánchez et al., 2021)** | | | | |
| task1_SINAI_TL_3.tsv_en (Best Model) | 0.777 | 0.775 | 0.781 | 0.774 |
| Baseline_svm_tfidf.tsv_en | 0.689 | 0.689 | 0.693 | 0.692 |
| Majority Class | 0.525 | 0.344 | 0.525 | 0.500 |

The results are averaged across 5 runs for each of the models for Task 1. The baseline and majority class results along with the results of the best model from the competition are presented as well.

TABLE III.
TASK 2 RESULTS

| Model | Accuracy | F1 Score | Precision | Recall |
|---|---|---|---|---|
| NBoW model | 0.474 | 0.108 | 0.166 | 0.082 |
| GLoVE embeddings | 0.471 | 0.107 | 0.165 | 0.082 |
| GLoVE embeddings (finetuned) | 0.473 | 0.108 | 0.166 | 0.082 |
| TFDistilBertForSequenceClassification | 0.420 | | | |
| DistilBERT NBoW (no dropout) | 0.582 | 0.486 | 0.497 | 0.502 |
| DistilBERT NBoW | 0.585 | 0.496 | 0.503 | 0.501 |
| DistilBERT + LSTM | 0.591 | 0.492 | 0.491 | 0.511 |
| DistilBERT + Bidirectional LSTM | 0.591 | 0.490 | 0.485 | 0.514 |
| DistilBERT + CNN | 0.605 | 0.502 | 0.491 | 0.529 |
| DistilBERT + MultiCNN | 0.603 | 0.498 | 0.486 | 0.527 |
| BERT + CNN | 0.607 | 0.499 | 0.482 | 0.531 |
| BERT + CNN + Dropout(0.4) | 0.597 | 0.494 | 0.484 | 0.514 |
| BERT + MultiCNN | 0.606 | 0.502 | 0.487 | 0.530 |
| BERT + MultiCNN + Dropout(0.4) | **0.610** | 0.508 | 0.499 | 0.530 |
| BERT + CNN + Data Augmentation | 0.609 | 0.514 | 0.507 | 0.529 |
| BERT + CNN + Data Augmentation (class weights) | 0.591 | 0.502 | 0.502 | 0.508 |
| **BERT + MultiCNN + Data Augmentation** | **0.610** | **0.519** | **0.514** | **0.531** |
| BERT + MultiCNN + Data Augmentation (class weights) | 0.595 | 0.511 | **0.515** | 0.515 |
| **Competition Results (Rodríguez-Sánchez et al., 2021)** | | | | |
| task2_LHZ_1.tsv_en (Best Model) | 0.634 | 0.560 | 0.551 | 0.574 |
| Baseline_svm_tfidf.tsv_en | 0.484 | 0.379 | 0.390 | 0.373 |
| Majority Class | 0.476 | 0.107 | 0.476 | 0.167 |

The results averaged across 5 runs for each of the models for Task 2. The baseline and majority class results along with the results of the best model from the competition are presented as well.

those using the GLoVE embeddings are seen to perform poorly compared to consequent models; with all three of them reaching accuracies below the baseline model. All the subsequent models which use either BERT or DistilBERT are seen to perform better than the baseline with similar accuracies to that of the best model from the competition. The best performance is seen with the BERT + MultiCNN models, shown in Fig. 1, with the best model having a dropout rate of 0.2 and achieving an average accuracy of 76.2%. Training the BERT + CNN and BERT + MultiCNN models with augmented training data did not improve performances for this task.

The results for Task 2, presented in Table III, are seen to be much lower than Task 1 as this was a multiclass classification task where the classes were no longer balanced. As before, the BoW and GLoVE models are seen to perform poorly, with these models seen to simply classify each text under the majority class of 'non-sexist'. Replacing these word embeddings with the DistilBERT and BERT variations shows significant improvements in the F1 scores for the test results. BERT achieves better performances than DistilBERT, with data augmentation improving these results both with and without using the class weights information. The best performance for this task is also seen with a BERT + MultiCNN model with this model using data augmentation where the operations of random insertion, synonym replacement, and random swap were applied at a rate of 0.05. However, this is not seen to be as competitive with the results produced by the best model (F1 score of 0.56) in the competition.

## VI. ERROR ANALYSIS

All the models using variations of BERT and DistilBERT are seen to improve results beyond the baselines reported through the competition, as well as perform similarly to the best results seen. However, all the models are still making mistakes. A manual analysis on the errors was done to understand the results and the challenges of this task, using the best models for the two tasks which are the BERT + MultiCNN model and the BERT + MultiCNN with Data Augmentation model for Task 1 and 2 respectively.

The confusion matrix for the BERT + MultiCNN model for Task 1 can be seen in Fig. 2. The numbers represent the ratio of classifications of the true label. This model is shown to be more likely to classify a text as sexist as opposed to non-

| Confusion Matrix for Task 1 | | |
|---|---|---|
| non-sexist | 0.69 | 0.31 |
| sexist | 0.15 | 0.85 |
| | non-sexist | sexist |
| | **Predicted Labels** | |

True Labels

Fig. 2. This shows the confusion matrix for the BERT + MultiCNN model's results for Task 1. The model is seen to be more likely to misclassify a text as sexist than non-sexist.

| Texts containing *'women, woman, girl, lady, female'* | | | No. of Tweets |
|---|---|---|---|
| **True Labels** non-sexist | 0.57 | 0.43 | 386 |
| sexist | 0.16 | 0.84 | 610 |
| | non-sexist | sexist | |
| | **Predicted Labels** | | |

| Texts containing *'feminis-'* | | | No. of Tweets |
|---|---|---|---|
| **True Labels** non-sexist | 0.04 | 0.96 | 23 |
| sexist | 0.03 | 0.97 | 140 |
| | non-sexist | sexist | |
| | **Predicted Labels** | | |

| Texts containing profanities[6] | | | No. of Tweets |
|---|---|---|---|
| **True Labels** non-sexist | 0.54 | 0.46 | 171 |
| sexist | 0.07 | 0.93 | 300 |
| | non-sexist | sexist | |
| | **Predicted Labels** | | |

Fig. 3. The confusion matrices showing the classification of tweets containing specific terms to see whether the model has biases in its classification.

sexist with 85% of the sexist tweets classified correctly compared to 69% of the non-sexist tweets.

A further analysis into the terms used in the texts was conducted, with these confusion matrices presented in Fig. 3. 69% of texts that contained the terms *'women, girl, female and lady'*, which are feminine terms, were labelled as 'sexist' while 31% were labelled as 'non-sexist'. However, the true annotations of these texts had 38% of these terms labelled as 'non-sexist'. When texts contained profanities[6], which are often used in social media casually, the model was seen to behave similarly with 76% of the texts labelled as sexist while the true labels had 64% of the texts labelled as such. When the texts contained the terms *'feminist(s)'* or *'feminism'*, the model was seen to classify a majority (97%) of the tweets as 'sexist' while only 86% of the tweets actually were annotated as 'sexist'.

The confusion matrix for the Task 2 results can be seen in Fig. 4. 'Sexual-violence' shows the best classification rate among the sexism categories, although this is still a low accuracy of 54%. Sexist texts are seen to be often misclassified under non-sexist, with this occurring more frequently for subtle categories of sexism, like ideological inequality or stereotyping dominance, as compared to categories which involve explicit sexism. This could be due to the fact that these categories of sexism are dependent on context and occur in more subtle and varied ways than, for example, sexual violence would be presented.

An example would be: *'woman will be driving and forget to stop at zebra crossings but they'll flash their lights to apologise so it's okay'* which was predicted as non-sexist but the true label was stereotyping dominance. This tweet utilises no profanities or explicit hatred but does convey a stereotype regarding women who drive. When further examining Task 2 predictions as done for Task 1, 65% of texts annotated as non-sexist which contained the terms *'feminist(s)'* or *'feminism'* were found to be predicted as ideological-inequality. However, this may have been due to fact that 80% of all texts containing these terms were annotated as ideological inequality in the dataset.

287 tweets (13%) were classified under a category of sexism when their true labels were 'non-sexist', and similarly, 250 texts (11.3%) were classified as 'non-sexist'

| Confusion Matrix for Task 2 | | | | | | |
|---|---|---|---|---|---|---|
| non-sexist | 0.73 | 0.04 | 0.03 | 0.07 | 0.09 | 0.05 |
| ideological-inequality | 0.24 | 0.47 | 0.01 | 0.08 | 0.17 | 0.03 |
| objectification | 0.19 | 0.03 | 0.35 | 0.11 | 0.19 | 0.13 |
| misogyny-non-sexual-violence | 0.21 | 0.05 | 0.04 | 0.51 | 0.13 | 0.06 |
| stereotyping-dominance | 0.24 | 0.05 | 0.06 | 0.10 | 0.53 | 0.02 |
| sexual-violence | 0.17 | 0.02 | 0.08 | 0.17 | 0.04 | 0.54 |
| | non-sexist | ideological-inequality | objectification | misogyny-non-sexual-violence | stereotyping-dominance | sexual-violence |
| | **Predicted Labels** | | | | | |

Fig. 4. The confusion matrix from the BERT + MultiCNN + Data Augmentation model for Task 2. The model is seen to misclassify many instances of sexism as 'non-sexist'.

---

[6] Profanities examined are: *'b\*tch, wh\*re, sk\*nk, f\*ck, sl\*t, c\*ck, c\*nt'*

TABLE IV.
MISCLASSIFICATION OF SEXIST TEXT IN TASK 2

| Misclassified as non-sexist | |
|---|---|
| stereotyping-dominance | 24.8% |
| objectification | 11.2% |
| misogyny-non-sexual-violence | 18.4% |
| ideological-inequality | 32.4% |
| sexual-violence | 13.2% |

This shows the percentages of sexist texts misclassified as 'non-sexist' with a breakdown of the different sexism labels they were annotated with. Fewer objectification and sexual violence annotated tweets are predicted to be non-sexist as opposed to ideological inequality and stereotyping dominance.

when they were annotated under one of the sexism categories. A breakdown of the latter misclassifications can be seen in Table IV. Categories like objectification or sexual violence where the sexism would present itself significantly less subtly is seen to be misclassified as 'non-sexist' at a lower rate than a category like ideological inequality which can present itself as a more covertly sexist statement as mentioned previously.

Some classifications predicted by the model appear to be more accurate than the true label, examples of which can be seen in Table V and VI. In Table V, the third example seems to be stating a fact as opposed to a sexist statement. Similarly, at first glance the last example seems to be correctly annotated but it may not be, as it is just a comment on feminism. Table VI presents similar examples for the multiclass classification task where the model appears to provide more accurate labels than in the annotated dataset. In

TABLE V.
SELECTED TASK 1 PREDICTIONS WHERE MODEL SEEMS MORE ACCURATE

| Preprocessed Text | Predicted Label | True Label |
|---|---|---|
| 'does anyone remember the evil mother hating sk*nk who disowned her own mother for being a trump supporter well guess what she still hates trump and she's still dribbling anti american diatribe like a typical liberal' | sexist | non-sexist |
| 'username i don't think feminists know what feminism is anymore and they always want to exclude the females who are more successful than them not feminists at all really' | sexist | non-sexist |
| 'kaliati says it is unfortunate that a day hardly passes without hearing of a case of rape defilement and other violence against girls women and children' | non-sexist | sexist |
| 'username of course you can be both i don't have a problem with radical feminism i don't agree with all of their views but i see what they want to achieve and i support it but that doesn't justify shitting on trans people again not all radfems do but some do and i'll call it out' | non-sexist | sexist |

This shows examples where the model seems to predict a more accurate label than the annotated one from the dataset for the binary classification task.

TABLE VI.
SELECTED TASK 2 PREDICTIONS WHERE MODEL SEEMS MORE ACCURATE

| Text | Predicted Label | True Label |
|---|---|---|
| 'username username i hope that is aimed at some women because i asked a woman if she wanted s*x after a few days i had known her and she accused me of sexual harassment it's a no win situation today' | sexual-violence | non-sexist |
| 'username i like netflix in an ironic twist of fate i'm watching it now something called bonding very good 17 minute episodes long enough to cover the important bits short enough to hold the attention alot like a **mini skirt**' | objectific-ation | non-sexist |
| 'leaders stop normalizing sexual harassment it's not okay do not call it fine or normal it is unacceptable' | non-sexist | sexual-violence |
| 'with internationalwomensday approaching we think the would be a safer place to live if we had more femaleleaders and a better place too with more women leaders in our communities and businesses womensrights womenempowerment womeninleadership womeninbusiness womensday ' | non-sexist | stereotyping-dominance |

This shows examples where the model seems to predict a more accurate label than the annotated one from the dataset for the binary classification task.

particular, these show when the model has chosen a sexism category while the annotated label is non-sexist or vice versa, with the last two examples again showing texts that are statements about women and sexual harassment as opposed to sexist opinions. With a dataset being annotated for something subjective like sexism, it is common for human coders to have differing opinions and consider sexist statements to be less offensive, or sexist, than they may actually be (Waseem and Hovy, 2016).

Table VII shows examples of incorrect predictions made by the model. The model is seen to often not recognise objectification in texts due to the varying forms in which it can be presented. Notably, the term *'mini-skirt'* occurs in an example in Table VI and VII (the term is in bold) with the model predicting one as objectification and the other as non-sexist.

TABLE VII.
SELECTED TASK 2 MODEL INCORRECT PREDICTIONS

| Text | Predicted Label | True Label |
|---|---|---|
| 'walks out on the tl with her very short **mini skirt**' | non-sexist | objectific-ation |
| 'username yassssss girl' | sexual-violence | non-sexist |
| 'sweet four attractive young ladies in dresses from chicks on the right discussing nra gun control on fox news this morning' | non-sexist | objectific-ation |

This shows examples where the model is seen to incorrectly predict labels for the multiclass classification task. In particular, objectification is seen to be hard for the model to identify.

There are also a few instances where both the model's prediction and the annotation seem inaccurate, potentially due to the casual use of profanities on social media. An example is: *"username b*tch how are you real you look like a princess"* which is predicted to be 'objectification' and has a true label of 'misogyny-non-sexual-violence', whereas this text seems to be a non-sexist comment. This example highlights the complexities of manually classifying sexist speech, which further shows how challenging this task is for a computer to accomplish.

Regarding the composition of the training and test datasets, although the training dataset consisted solely of tweets while the test set contained both tweets and gabs, a large discrepancy was not seen between accuracies for the two types of texts. 77% of both the tweets and gabs in the test dataset were classified correctly for task 1, whereas for task 2, the correct classification rate was 61% for tweets and 58% for gabs. This could indicate that the length of texts, with a maximum of 280 characters on Twitter[7] compared to 3000 characters per post on Gab[8], did not affect the model's performance, potentially due to the use of the CNN layer in the model.

The specific effect of length of text on the models performances was looked at further in Table VIII. There are some variations seen in the accuracies for different lengths of texts, with the model seen to be more accurate with the longer (500+) and shorter (<100) texts. This could be attributed to potentially having more context with longer texts, and fewer features to predict on with the shorter ones, with more room for confusion with texts of length 100 to 500 characters. However, further experimentation with balanced datasets would be required to determine this.

### VII. CONCLUSION AND FUTURE WORK

The increased use of social media has enabled hate speech, including sexist speech, to easily propagate and affect people globally. With online hate speech linked to offline violence, it is essential to successfully classify speech as hateful through automatic methods. This paper presented a variety of deep neural networks using BERT and DistilBERT to differentiate sexist tweets and gabs from non-sexist ones, as well as further classify sexist text into types of sexism using the EXIST dataset. The best model for the binary classification task used BERT along with a CNN architecture using filter sizes of 4, 6, and 8 to achieve an accuracy of 76.2% while the best accuracy from the competition was 77%. The same model along with data augmentation achieved the best performance on the multi-class classification task with an F1 score of 51.9% which was lower than the best F1 score (56%) from the competition.

Due to the subjectivity involved in annotating sexist text as well as the complexity of natural language in tweets and gabs, this task proved to be challenging for the models to achieve ideal results on. Categories which contain explicit hatred like 'sexual-violence' were seen to be labelled more accurately than more subtle instances of sexism such as those seen under 'ideological-inequality'. Profanities are often used on social media platforms without necessarily insinuating sexist speech, such as in casual conversation or song lyrics and hence were not very helpful features for classification. Similarly, speech can be sexist without using any explicit and specific words to indicate this. Due to the varying perceptions of sexism by humans, some labels within the dataset were found to have been potentially mislabelled, which creates further challenges for using the dataset to train models.

The type of texts (tweets versus gabs) was not seen to affect the BERT + MultiCNN model's performance significantly for either task. This could be attributed to the use of the convolutional layer with different filter sizes with max pooling. The model was seen to perform better for text of length 500+ characters and <100 characters, although the impact of length would need to be further examined to confirm this by creating more balanced datasets for this purpose.

Another avenue for further exploration could be using additional features such as the gender or ethnicity of authors as stated by Waseem and Hovy (2016), although this information may be challenging to obtain. The models trained on the EXIST dataset could also be tested on a different sexism dataset to observe the generalisability of the models across different annotated data. Finally, a benchmark annotated dataset for sexism would also allow for better development and comparison of models.

TABLE VIII.
BREAKDOWN OF CORRECT PREDICTIONS IN TASK 1 AND TASK 2 BY TEXT LENGTH

| No. of Characters in the Text | Total No. of Texts (Training Dataset) | Total No. of Texts (Test Dataset) | Percentage Correctly Predicted | |
| --- | --- | --- | --- | --- |
| | | | Task 1 | Task 2 |
| 0-100 | 1123 | 683 | 79.6% | 66.2% |
| 101-250 | 1534 | 965 | 74.8% | 59.4% |
| 251-500 | 762 | 536 | 76.7% | 53.0% |
| 501-1000 | 17 | 14 | 92.9% | 71.4% |
| 1001+ | 0 | 10 | 80.0% | 80.0% |

This shows the percentage of correct predictions by the best models for each task, further split into the ranges of text lengths. Most texts are seen to be between the lengths of 0-250 characters.

---

[7] https://help.twitter.com/en/using-twitter/how-to-tweet

[8] https://help.gab.com/article/basics-post-composer-options

REFERENCES

Ahluwalia, R., Shcherbinina, E., Callow, E., Nascimento, A., and Cock, M. D., 2018. Detecting Misogynous Tweets. *IberEval@SEPLN*.

Alammar, J., 2018. *The Illustrated BERT, ELMo, and co. (How NLP Cracked Transfer Learning)*. [Online] Available at: https://jalammar.github.io/illustrated-bert/ [Accessed 30 June 2021].

Brown, A., 2017. What is hate speech? Part 1: The Myth of Hate. *Law and Philosophy,* 36(4), pp. 419-168.

Davidson, T., Warmsley, D., Macy, M. and Weber, I., 2017. Automated Hate Speech Detection and the Problem of Offensive Language. *ICWSM*.

Devlin, J., Chang, M. W., Lee, K. and Toutanova, K., 2018. BERT: Pre-training of Deep Bidirectional Transformers for Language Understanding. *arXiv preprint arXiv:1810.04805*.

Djuric, N., Zhou, J., Morris, R., Grbovic, M., Radosavljevic, V., and Bhamidipati, N., 2015. Hate Speech Detection with Comment Embeddings. *Proceedings of the 24th International Conference on World Wide Web,* pp. 29-30.

Fersini, E., Rosso, P. and Anzovino, M., 2018. Overview of the Task on Automatic Misogyny Identification at IberEval 2018. *IberEval@SEPLN*.

Gianfortoni, P., Adamson, D. and Rosé, C. P., 2011. Modeling of Stylistic Variation in Social Media with Stretchy Patterns. *Proceedings of the First Workshop on Algorithms and Resources for Modelling of Dialects and Language Varieties,* pp. 49-59.

Goenaga, I., Atutxa, A., Gojenola, K., Casillas, A., Ilarraza, A. D., Ezeiza, N., Oronoz, M., Pérez, A. and Perez-de-Viñaspre, O., 2018. Automatic Misogyny Identification Using Neural Networks. *IberEval@SEPLN*.

Hatzipanagos, R., 2018. *How online hate turns into real-life violence*. [Online] Available at: https://www.washingtonpost.com/nation/2018/11/30/how-online-hate-speech-is-fueling-real-life-violence/ [Accessed 17 June 2021].

Kwok, I. and Wang, Y., 2013. Locate the Hate: Detecting Tweets against Blacks. *AAAI*.

Laub, Z., 2019. *Hate Speech on Social Media: Global Comparisons*. [Online] Available at: https://www.cfr.org/backgrounder/hate-speech-social-media-global-comparisons [Accessed 17 June 2021].

Mandl, T., Modha, S., Majumder, P., Patel, D., Dave, M., Mandlia, C., and Patel, A., 2019. *Overview of the HASOC Track at FIRE 2019: Hate Speech and Offensive Content Identification in Indo-European Languages.* New York, s.n.

Mehdad, Y. and Tetreault, J. R., 2016. Do Characters Abuse More Than Words?. *SIGDIAL Conference*.

Mondal, M., Silva, L. A. and Benevenuto, F., 2017. A Measurement Study of Hate Speech in Social Media. *Proceedings of the 28th ACM Conference on Hypertext and Social Media,* p. 85–94.

Pamungkas, E. W., Cignarella, A. T., Basile, V. and Patti, V., 2018. Automatic Identification of Misogyny in English and Italian Tweets at EVALITA 2018 with a Multilingual Hate Lexicon. *EVALITA@CLiC-it*.

Parikh, P., Abburi, H., Badjatiya, P., Krishnan, R., Chhaya, N., Gupta, M., and Varma, V., 2019. Multi-label Categorization of Accounts of Sexism using a Neural Network. *arXiv preprint arXiv:1910.04602*.

Pennington, J., Socher, R. and Manning, C., 2014. *Proceedings of the 2014 Conference on Empirical Methods in Natural Language Processing ({EMNLP}).* Doha, s.n.

Qian, J., Bethke, A., Liu, Y., Belding, E., and Wang, W. Y., 2019. A Benchmark Dataset for Learning to Intervene in Online Hate Speech. *arXiv preprint arXiv:1909.04251*.

Raiyani, K., Gonçalves, T., Quaresma, P. and Nogueira, V. B., 2018. Fully Connected Neural Network with Advance Preprocessor to Identify Aggression over Facebook and Twitter. *Proceedings of the first workshop on trolling, aggression and cyberbullying (TRAC-2018),* pp. 28-41.

Relia, K., Li, Z., Cook, S. H. and Chunara, R., 2019. Race, Ethnicity and National Origin-Based Discrimination in Social Media and Hate Crimes across 100 U.S. Cities. *Proceedings of the International AAAI Conference on Web and Social Media,* Volume 13, pp. 417-427.

Rodríguez-Sánchez, F., Carrillo-de-Albornoz, J. and Plaza, L., 2020. Automatic Classification of Sexism in Social Networks: An Empirical Study on Twitter Data. *IEEE Access,* Volume 8, pp. 219563-219576.

Rodríguez-Sánchez, F., Carrillo-de-Albornoz, J., Plaza, L., Gonzalo, J., Rosso, P., Comet, M., and Donoso, T., 2021. Overview of EXIST 2021: sEXism Identification in Social neTworks. *Procesamiento del Lenguaje Natural,* Volume 67.

Samghabadi, N. S., Patwa, P., Srinivas, P., Mukherjee, P., Das, A., and Solorio, T., 2020. Aggression and Misogyny Detection using BERT: A Multi-Task Approach. *Proceedings of the Second Workshop on Trolling, Aggression and Cyberbullying,* pp. 126-131.

Sanh, V., Debut, L. C. J. and Wolf, T., 2019. DistilBERT, a distilled version of BERT: smaller, faster, cheaper and lighter. *ArXiv,* Volume abs/1910.01108.

Shushkevich, E. and Cardiff, J., 2018. Misogyny Detection and Classification in English Tweets: The Experience of the ITT Team. *EVALITA@CLiC-it*.

Suler, J., 2004. The Online Disinhibition Effect. *Cyberpsychology and Behavior,* 7(3), pp. 321-326.

UN Women, 2020. *Online and ICT-facilitated violence against women and girls during COVID-19.* s.l.:United Nations Entity for Gender Equality and the Empowerment of Women (UN Women).

Vigna, F., Cimino, A., Dell'Orletta, F., Petrocchi, M., and Tesconi, M., 2017. *Hate Me, Hate Me Not: Hate Speech Detection on Facebook.* s.l., s.n.

Vogels, E., 2021. *The State of Online Harassment.* [Online] Available at: https://www.pewresearch.org/internet/2021/01/13/the-state-of-online-harassment/ [Accessed 6 June 2021].

Waseem, Z. and Hovy, D., 2016. Hateful Symbols or Hateful People? Predictive Features for Hate Speech Detection on Twitter. *Proceedings of the NAACL Student Research Workshop,* pp. 88-93.

Wright, M. F., Harper, B. and Wachs, S., 2019. The associations between cyberbullying and callous-unemotional traits among adolescents: The moderating effect of online disinhibition. *Personality and Individual Differences,* Volume 140, pp. 41-45.

Xiang, G., Fan, B., Wang, L., Hong, J., and Rose, C., 2012. Detecting Offensive Tweets via Topical Feature Discovery Over a Large Scale Twitter Corpus. *Proceedings of the 21st ACM International Conference on Information and Knowledge Management,* pp. 1980-1984.

Zhang, Z. and Luo, L., 2019. Hate Speech Detection: A Solved Problem? The Challenging Case of Long Tail on Twitter. *Semantic Web,* Volume 10, pp. 925-945.

Zou, K. and Wei, J., 2019. EDA: Easy Data Augmentation Techniques for Boosting Performance on Text Classification Tasks. *arXiv preprint arXiv:1901.11196.*